\documentclass{article}
\usepackage{iclr2026_conference,times}
\usepackage{float}   
\iclrfinalcopy

\usepackage{amsmath,amsfonts,bm}

\def\eqref#1{equation~\ref{#1}}

\def\1{\bm{1}}

\def\rt{{\textnormal{t}}}

\DeclareMathAlphabet{\mathsfit}{\encodingdefault}{\sfdefault}{m}{sl}
\SetMathAlphabet{\mathsfit}{bold}{\encodingdefault}{\sfdefault}{bx}{n}

\usepackage{hyperref}
\hypersetup{colorlinks=true, citecolor=blue, linkcolor=blue, urlcolor=blue}
\usepackage{url}
\usepackage{booktabs}
\usepackage{array}
\usepackage{amsmath}
\usepackage{amssymb}
\usepackage{graphicx}
\usepackage{microtype}
\usepackage{enumitem}
\setlist{nosep, leftmargin=1.4em}
\newcommand{\rto}{\mathrm{ratio\_true}}
\newcommand{\slope}{\mathrm{slope}}
\newcommand{\eff}{\mathrm{eff}}
\newcommand{\hit}{\mathrm{hit}}
\newcommand{\Corr}{\mathrm{Corr}}

\title{Evidence-Type Competition: When Can\\ Interventional Data Teach Language\\ Models Causal Direction?}

\author{Xining Xun \\
Tsingjiao Information Science (Beijing) Co., Ltd.
}

\begin{document}

\maketitle
\lhead{Preprint}

\begin{abstract}
Interventional data is widely regarded as the gold standard for teaching models causal reasoning. We test this assumption in a fully controlled synthetic environment pitting observational correlation against causal effect, and find it fails instructively. In Simpson's-paradox worlds---where the two have systematically opposite signs---increasing the fraction of interventional samples in pretraining ($\alpha$) does \emph{not} improve causal direction: the magnitude of the model's $do()$-response grows monotonically with $\alpha$, yet its sign is copied from the observational context. What governs whether interventional evidence is used is not the training mixture but the \emph{evidence type present in the context at inference time}. Under an identical training recipe, a purely observational context induces systematic sign reversal in 29/50 worlds, a mixed context in 19/50, while aligned interventional probes alone yield 41/50 correct, with reversals at the measurement noise floor. Inference-time interventions localize the phenomenon: doubling the probe dose does not lift the suppression, whereas \emph{erasing observational evidence from the context immediately releases the suppressed causal interpolation ability} ($\rto = +0.56$, matching a model trained on probes alone); a four-state content manipulation shows the switch is content-mediated and graded. The suppression is stable across training seeds---every measurable strong reversal (11/11) persists on a matched-protocol second seed---and robust as a rate at 0.93B parameters (31.8\% vs.\ 6\% reversals in the matched probe-only arm), even as absolute interventional gains shrink four-fold. An external audit on CLadder further exposes a learned positive-effect prior with a two-layer structure: sign-randomized retraining removes it in-distribution but not out-of-distribution, where a positive default persists. We summarize: \emph{the capability lives in the weights; the switch lives in the context}---activation patching localizes the switch to the middle layers' observational rows. We further quantify the sampling noise floor of probe-based causal evaluation and an evidence-averaging protocol that cuts sign errors from 26\% to 9\%.
\end{abstract}

\vspace{0.2ex}

\section{Introduction}
\label{sec:intro}

\textbf{The interventional-data hypothesis.} The ladder of causation \citep{pearl2009causality} places intervention above observation: identifying causal effects from correlation requires, in principle, data from interventional distributions. This hierarchy has shaped data-curation intuitions in the LLM era. A natural hypothesis---which we denote \textbf{H1}---holds that \emph{increasing the proportion of interventional data in pretraining should improve a model's causal reasoning}. H1 is implicit in much data-mixture practice, yet to our knowledge has never been tested under fully controlled conditions: known confounding structure, computable causal ground truth, a leak-free protocol.

\textbf{A negative result that is more informative than ``false''.} We test H1 on a family of Simpson's-paradox worlds in which $\Corr_{obs}(X,Y) < 0$ while the true causal effect $\eff(X \to Y) > 0$, with opposite signs by construction. A six-level sweep of the interventional fraction $\alpha \in \{0, 0.1, \dots, 1.0\}$ shows no monotone trend in corrected sign accuracy (0.5/0.3/0.2/0.4/0.4/0.4): \textbf{H1 is falsified}. The manner of failure contains the real discovery: the \emph{magnitude} of the model's Simpson-slope grows monotonically with $\alpha$ (0.169 $\to$ 0.384), while its \emph{sign} stays wrong throughout. The model learns something from interventional data---the magnitude---but copies the direction from the observational context: a \textbf{magnitude--direction duality}.

\textbf{The evidence-type factor.} If the mixture ratio is not the operative variable, what is? We isolate a one-dimensional factor---the \emph{evidence type in the context}---holding training fixed at $\alpha = 1.0$. Under a 50-world, evidence-averaged protocol, the four-way comparison (Table~\ref{tab:main}) reveals a \textbf{contextual dose gradient}: the more observational evidence in the context, the stronger the suppression of causal direction (mean $\rto$: pure obs $-0.288$ $\to$ mixed $-0.092$ $\to$ pure probes $+0.418$; paired McNemar $p$ down to $1.1\mathrm{e}{-9}$).

\begin{figure}[!t]
\centering
\includegraphics[width=\textwidth]{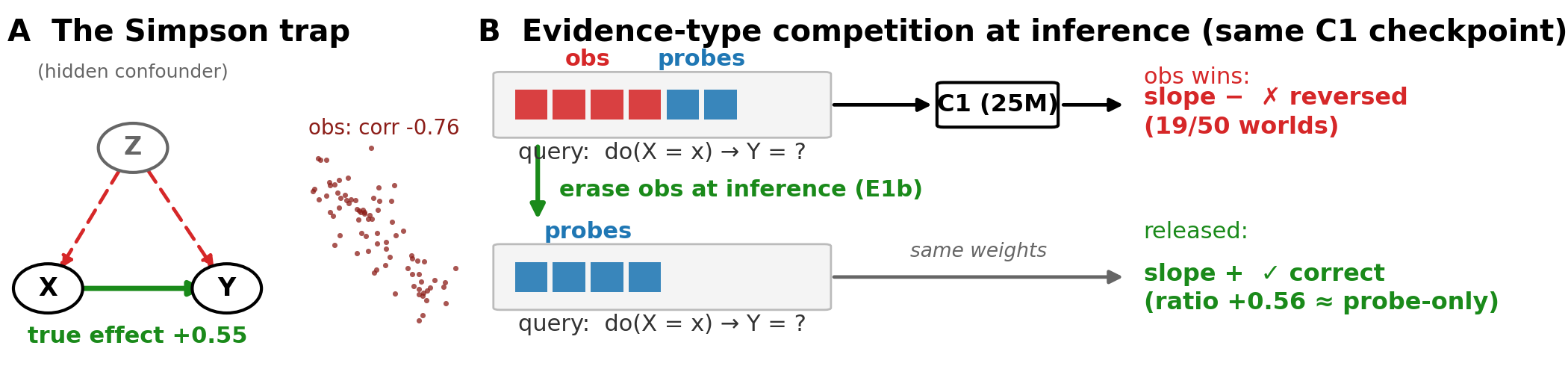}
\caption{Overview (schematic). \textbf{A:} a hidden confounder makes observational correlation negative while the true effect of $X$ on $Y$ is positive (sampled world, $\rho = 0.9$; full tour in Appendix~\ref{app:construction}). \textbf{B:} with observational records in context, the answer follows the observational sign (19/50 reversed); erasing them \emph{at inference time} releases the suppressed interpolation (E1b: $\rto = +0.56$); doubling the probe dose alone does not (E1a).}
\label{fig:teaser}
\end{figure}

\textbf{Mechanism: the switch is in the context} (Figure~\ref{fig:teaser}). Three zero-training-cost, inference-time interventions (E1) decide whether the suppression lives in training or at inference time. (i) Doubling the probe dose does not lift it. (ii) \emph{Erasing the observational evidence immediately releases the suppressed interpolation ability} ($\rto$ rises from $-0.09$ to $+0.56$, matching the probe-only model). (iii) Dose titration and content manipulation (E5) show the switch is \emph{content-mediated and graded}: one observational record barely suppresses, two dilute, four capture---\emph{first dilution, then capture}---and removing observational \emph{content} releases suppression even when observational-\emph{format} tokens remain. The interpolation circuit was in the weights all along; the context had switched it off.

\textbf{Contributions.}
\begin{enumerate}
\item \textbf{A motivating negative result.} H1 is falsified under controlled conditions, revealing the magnitude--direction duality (\S\ref{sec:h1}).
\item \textbf{Main result.} The evidence-type factor and the contextual dose gradient: a 50-world four-arm comparison with paired statistics (\S\ref{sec:main}, Table~\ref{tab:main}).
\item \textbf{Mechanism contribution.} Inference-time decision matrices (E1, E5) localizing suppression as an inference-time, content-mediated, graded phenomenon (\S\ref{sec:e1}, \S\ref{sec:e5}).
\item \textbf{Methodology.} Quantification of the sampling noise floor of probe-based causal evaluation: a perfect interpolator commits 26\% sign errors single-shot; evidence averaging repairs this (\S\ref{sec:method}).
\item \textbf{Robustness and boundary conditions.} Rate-level persistence of the suppression at 0.93B scale (\S\ref{sec:e6}); a CLadder audit exposing a learned positive-effect prior and its two-layer structure under sign randomization (\S\ref{sec:e7}, \S\ref{sec:e8}).
\end{enumerate}

\section{Related Work}
\label{sec:related}

\textbf{Causal reasoning of LLMs.} CLadder \citep{jin2023cladder} systematically evaluates LLMs across the three rungs of the causal ladder, finding degraded accuracy on formal causal queries; Corr2Cause \citep{jin2024corr2cause} finds near-chance performance at inferring causation from correlational statements; CausalPitfalls \citep{du2025causalpitfalls} audits statistical pitfalls including Simpson's paradox; \citet{kiciman2023causal} report strong performance of GPT-family models on causal commonsense tasks. Unlike these studies, which \emph{evaluate existing models}, we \emph{control the training distribution itself}.

\textbf{Amortized causal inference.} A complementary line pretrains transformers on synthetic causal data to estimate effects in-context: Do-PFN \citep{robertson2025dopfn} predicts interventional outcomes from observational context with prior-fitted networks; CausalPFN \citep{balazadeh2025causalpfn} amortizes ATE estimation under ignorability; \citet{liang2025endogeneity} show that transformers emulate 2SLS when facing endogeneity in in-context linear regression. These works ask \emph{whether} a transformer can amortize causal estimation; we ask \emph{which evidence type} in a mixed observational--interventional pretraining stream drives the interventional answer, and what arbiters---dose, content, and sign priors---decide the competition.

\textbf{Teaching causal competence through training data.} Closest to our interventional stance, \citet{vashishtha2024axiomatic} train transformers from scratch on symbolic demonstrations of causal axioms (transitivity, $d$-separation) and show generalization to more complex graph structures, while \citet{lampinen2023passive} show that agents can acquire active causal strategies from purely passive data---provided they may \emph{intervene at test time}, with natural-language explanations enabling generalization even from perfectly confounded training data. Neither, however, mixes competing evidence types: the first teaches explicit symbolic rules; the second assumes an active experimentation channel at deployment. We isolate the corner they leave open---purely passive inference over contexts where observational and interventional evidence coexist and contradict---and find that an observational prior, not missing capability, decides the answer (\S\ref{sec:main}, \S\ref{sec:e5}).

\textbf{In-context learning as function learning.} \citet{garg2022what} show that Transformers can learn linear and nonlinear function classes from in-context examples (our least-squares-interpolator baseline is of this kind); \citet{akyurek2023what} relate ICL to gradient-descent-like algorithms. Our probe protocol is the causal version of this paradigm: probes supply $(X, \text{responses of other variables})$ under intervention, and the query asks for interpolation over a $do(X)$ grid. Our negative result also delineates a boundary for this paradigm: when observational evidence is mixed into the context, ICL selects the wrong function class---not a failure to learn, but a failure of evidence selection.

\textbf{Knowledge conflicts.} \citet{longpre2021entity} and \citet{xie2024adaptive}, among others, study LLM behavior when parametric knowledge conflicts with contextual knowledge, finding that reliance on context varies with prior strength. Ours is a sharper causal instance: the conflicting parties are two \emph{statistical evidence types} rather than factual memories. To our knowledge, the specific form ``interventional evidence suppressed at inference time by an observational prior'' has not been reported.

\textbf{Simpson's paradox.} A classical statistical problem \citep{pearl2009causality}; in NLP it has appeared mainly in bias auditing. We repurpose it as a jointly controlled train--evaluation instrument: a world family in which observational correlation and causal effect have opposite signs makes ``copying observation'' and ``true causation'' mechanically distinguishable at the level of signs.

\section{Setup}
\label{sec:setup}

\textbf{Synthetic world generator.} Each world is a parametric structural causal model (SCM): $n$ nodes ($n \in [5, 12]$; the novel-topology domain uses $[13, 16]$), Erd\H{o}s--R\'enyi or scale-free graph structure (equal probability), and mixed linear--nonlinear mechanisms. A confounding-strength parameter $\rho \in \{0, .3, .6, .9\}$ controls unobserved confounding paths ($Z \to X$, $Z \to Y$). The evaluation family (SEED\_TEST domain) uses $\rho = 0.9$ and selects confounded triples $(Z, X, Y)$ such that observational correlation and causal effect have opposite signs (a \emph{Simpson trap}): in 50/50 worlds, $\Corr_{obs}(X,Y) < 0$ while $\eff(X \to Y) > 0$ (mean $\approx +0.57$; a sampled world is visualized end to end in Figure~\ref{fig:construction}, Appendix~\ref{app:construction}). Training data are generated online and are infinitely fresh, eliminating fixed-corpus memorization.

\textbf{Evidence formats.} Each training/evaluation sample = a context ($k = 4$ evidence records) + one query. Five evidence types: \textbf{\textsc{obs}} (observational samples); \textbf{\textsc{none}} (no context); \textbf{\textsc{int}} (intervention records on \emph{random} nodes); \textbf{\textsc{probe}} (intervention records on the \emph{query} node $X$---$do(X{=}v) \to$ responses of all variables, $v$ sampled 50/50 from a two-sided range); \textbf{\textsc{obs\_probe}} (half obs, half probes). Queries take the form $do(X{=}x_0) \to Y{=}?$, over the grid $x_0 \in \mathcal{X} = \{-1, -0.5, 0, 0.5, 1\}$.

\textbf{Model and training.} A 25.7M-parameter GPT-style decoder (8 layers, na\"ive attention), vocabulary 41, block size 640. Training: 20{,}000 steps, batch 64, lr $6\mathrm{e}{-4}$ (warmup 500). $\alpha$-mixture sampling: each sample is independently drawn from the interventional distribution with probability $\alpha$ (strict binomial mixture). Seed domains are disjoint across train/val/test/novel.

\textbf{Formal statement.} A world is an SCM $W = (\mathcal{G}, \mathcal{F}, P_U)$ over variables $V = \{V_1, \dots, V_n\}$ with mechanisms $V_i = f_i(\mathrm{pa}(V_i), U_i)$ and an unobserved confounder $Z$ for a designated triple $(Z, X, Y)$, with the Simpson-trap condition enforced in the evaluation family:
\begin{equation}\label{eq:trap}
\Corr_{obs}(X, Y) \cdot \eff(X \to Y) < 0,
\qquad
\eff(X \to Y) = \tfrac{1}{2}\big( \mathbb{E}[Y \mid do(X{=}1)] - \mathbb{E}[Y \mid do(X{=}-1)] \big) .
\end{equation}
A context is a multiset of $k$ evidence records $C = \{r_1, \dots, r_k\}$, each either \emph{observational} $r^{obs} = (v_1, \dots, v_n) \sim P_W(V)$ or \emph{interventional}
\begin{equation}\label{eq:rec}
r^{int}_j = \big(do(V_j{=}v),\; (v_1, \dots, v_n) \sim P_W(V \mid do(V_j{=}v))\big) .
\end{equation}
Evidence type $\tau(C) \in \{\textsc{obs}, \textsc{none}, \textsc{int}, \textsc{probe}, \textsc{obs\_probe}\}$ fixes which records appear. Per world $w$, predictions $\hat{Y}_{x_0}$ over the grid are regressed by least squares,
\begin{equation}\label{eq:slope}
\slope_w = \arg\min_{b}\;\min_{a} \sum_{x_0 \in \mathcal{X}} \big(\hat{Y}_{x_0} - (a + b\, x_0)\big)^2,
\qquad
\rt_w = \slope_w \,/\, \slope\_\mathrm{true}_w ,
\end{equation}
where $\slope\_\mathrm{true}_w$ is the oracle grid-regression slope from $\mathbb{E}[Y \mid do(X{=}x_0)]$ estimated by large-sample Monte Carlo. $\rto \approx 1$: correct magnitude; $\approx 0$: blind; $<0$: reversed. The hit criterion applies a magnitude gate $\tau_0 = 0.05$:
\begin{equation}\label{eq:hit}
\hit_w = \mathbb{1}\big[\,|\slope_w| > \tau_0 \;\wedge\; \mathrm{sign}(\slope_w) = \mathrm{sign}(\eff_w)\,\big] .
\end{equation}
We also track the interventional point-prediction advantage $\Delta = \mathrm{mse}_{obs} - \mathrm{mse}_{do}$ over held-out $do()$-queries ($\Delta > 0$: interventional semantics learned).

\section{Evaluation Methodology}
\label{sec:method}

\textbf{Hit criterion.} Per-world three-way classification: \textbf{hit} ($|\slope| > 0.05$, sign consistent with $\eff$) / \textbf{flat} ($|\slope| \le 0.05$) / \textbf{reversed} ($\slope < -0.05$; $\eff$ is always positive in the evaluation family). The magnitude gate eliminates false positives in which a numerically flat line is counted as a hit.

\textbf{The sampling noise floor of probe-based evaluation.} Even if the model were a \emph{perfect least-squares interpolator}, probe evidence is itself a single-draw random variable (4 probes, one sample each), and its regression slope carries an intrinsic sign-error rate. Writing the oracle grid response as $\mu(x_0) = \mathbb{E}[Y \mid do(X{=}x_0)]$, single-shot evaluation estimates $\tilde{\mu}(x_0) = \mu(x_0) + \varepsilon(x_0)$ with heteroscedastic $\varepsilon$ inherited from $P_W$, so that
\begin{equation}\label{eq:floor}
p_{\mathrm{err}} = \Pr\big[\mathrm{sign}\big(\slope(\tilde{\mu})\big) \neq \mathrm{sign}\big(\slope(\mu)\big)\big] > 0
\quad \textit{even for a perfect model.}
\end{equation}
Offline simulation (20 repetitions): under single-shot sampling ($n_{rep}{=}1$), a perfect interpolator commits sign errors at a rate of \textbf{2.6/10 (26\%)}---per-world sign statistics that do not control for this floor can manufacture ``attractor camps'' out of noise. The repair is \textbf{evidence averaging} ($n_{rep}$): for each query, draw $n_{rep}$ independent context-evidence sets, average the predictions, then regress, reducing $\mathrm{Var}[\tilde{\mu}]$ by a factor $\approx 1/n_{rep}$; $n_{rep} = 8$ reduces the floor to \textbf{0.9/10 (9\%)}. All final results are obtained at $n_{rep} = 8$; reversal counts are tested one-sidedly against the floor,
\begin{equation}\label{eq:binom}
\Pr\big[\mathrm{Bin}(50,\; p_0{=}0.09) \geq k\big],
\end{equation}
to separate mechanistic signal from measurement noise (Figure~\ref{fig:floor}).

\textbf{Statistics.} Arms are compared on identical world sets: exact McNemar test on hit discordance, Wilcoxon signed-rank on $\rto$, and the binomial test of \eqref{eq:binom}.

\textbf{Reproducibility.} The generator's intervention-value cache persists across calls: offline replay of the evaluation stream must explicitly re-seed the cache, otherwise identical protocols drift. Code fingerprints (md5) and seeds for all final results are registered in Appendix~\ref{app:repro}.

\section{Results}
\label{sec:results}

\subsection{H1 falsified: the magnitude--direction duality}
\label{sec:h1}

Fixing $\rho = 0.9$ with \textsc{obs} contexts, we sweep $\alpha \in \{0.0, 0.1, 0.3, 0.5, 0.7, 1.0\}$ (20{,}000 steps each). \textbf{H1 is falsified}: corrected sign accuracy across the six levels is 0.5/0.3/0.2/0.4/0.4/0.4---no monotone trend (the 0.5 at $\alpha = 0$ is blind sign-hitting). \textbf{Duality}: the \emph{magnitude} of the Simpson slope grows monotonically with $\alpha$ (0.169 $\to$ 0.384) while the \emph{sign} stays negative throughout: \emph{magnitude is learned from interventions; direction is copied from context}. \textbf{Loss decouples from causal competence}: all six runs converge with similar losses, yet causal metrics diverge. \textbf{The shortcut deepens with training}: at $\alpha = 1.0$, $\Delta$ decays from $+0.20$ (step 2k) to $-0.19$ (step 20k)---interventional semantics is learned first, then overwritten (replayed within a single run, \S\ref{sec:robust}).

\textbf{The dual role of context (ablations).} Training \emph{without} context (\textsc{none}) removes the contamination source (reversals 0/10) but collapses the slope magnitude to 0.018---the model loses its world-identification channel and falls near-blind. Random probes (\textsc{int}) are worse than baseline (reversals 7/10), rejecting the ``target-node confusion'' hypothesis. Context is simultaneously a \emph{contamination source} and a \emph{world-identification channel}.

\subsection{Main result: the four-arm evidence-type comparison}
\label{sec:main}

\begin{table}[htbp]
\caption{Four-arm comparison over 50 worlds ($\alpha = 1.0$, $\rho = 0.9$, $n_{rep} = 8$, identical world set, paired; curr.\ = curriculum).}
\label{tab:main}
\centering\small
\begin{tabular}{llccccc}
\toprule
Arm & Context evidence & Hit & Flat & Reversed & mean $\rto$ & $\Delta$ \\
\midrule
Baseline & pure \textsc{obs} & 7/50 & 14 & 29 & $-0.288$ & $-0.100$ \\
C1 & \textsc{obs} + probes & 18/50 & 13 & 19 & $-0.092$ & $+0.041$ \\
\textbf{C2} & pure probes & \textbf{41/50} & 5 & 4 & $\mathbf{+0.418}$ & $+0.262$ \\
C2-curr2 & pure probes (harder curr.) & 40/50 & 6 & 4 & $+0.482$ & $+0.373$ \\
\bottomrule
\end{tabular}
\end{table}

\begin{figure}[htbp]
\centering
\includegraphics[width=0.50\textwidth]{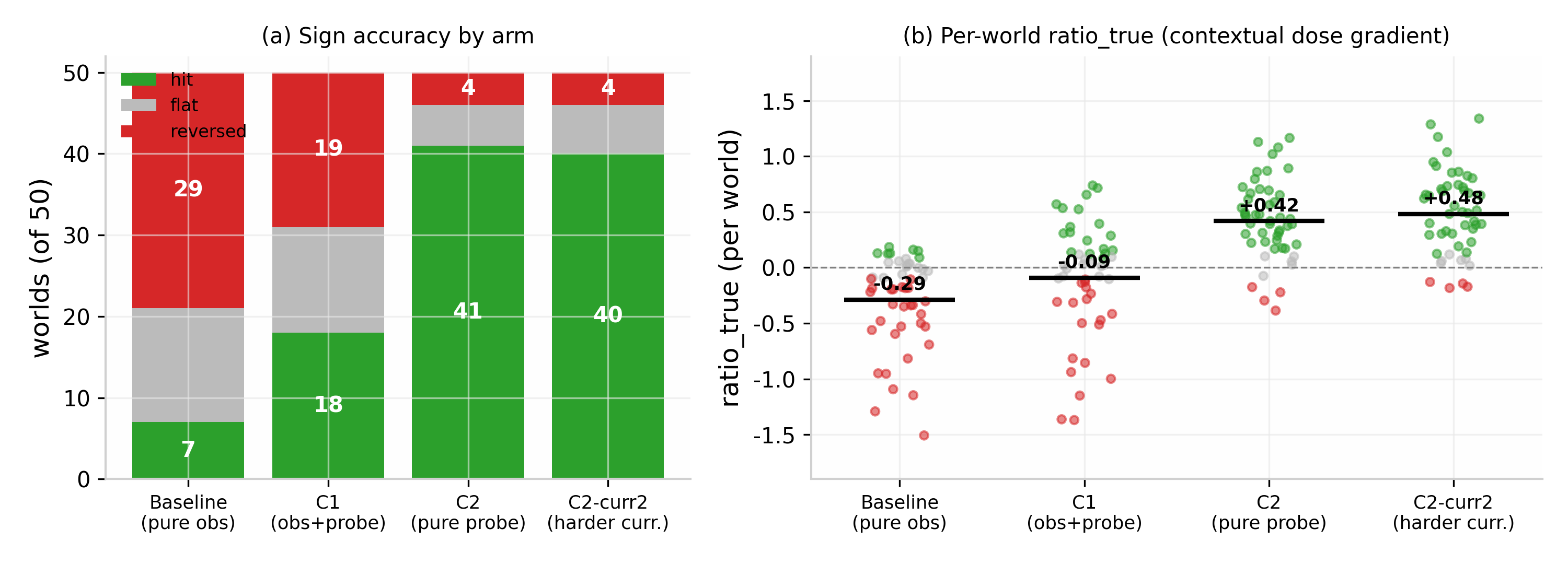}
\caption{Main result. (a) Hit/flat/reversed counts over 50 worlds; (b) per-world $\rto$ distributions---the contextual dose gradient.}
\label{fig:main}
\end{figure}

Table~\ref{tab:main} and Figure~\ref{fig:main} show the \textbf{contextual dose gradient}: as the observational dose in the context decreases (pure obs $\to$ mixed $\to$ pure probes), mean $\rto$ rises monotonically ($-0.288 \to -0.092 \to +0.418$), with $\Delta$ in lockstep ($-0.100 \to +0.041 \to +0.262$). Paired exact McNemar tests (hit criterion): C2 vs.\ baseline $b{=}35/c{=}1$, $p = 1.1\mathrm{e}{-9}$; C2 vs.\ C1 $b{=}25/c{=}2$, $p = 5.7\mathrm{e}{-6}$; C1 vs.\ baseline $b{=}14/c{=}3$, $p = 1.3\mathrm{e}{-2}$. Paired Wilcoxon on $\rto$: $p = 7.6\mathrm{e}{-9}$ / $2.9\mathrm{e}{-7}$ / $5.8\mathrm{e}{-3}$. C2's slope exceeds C1's in 43/50 worlds (paired $t$, $p = 5.6\mathrm{e}{-8}$).

\begin{figure}[htbp]
\centering
\includegraphics[width=0.40\textwidth]{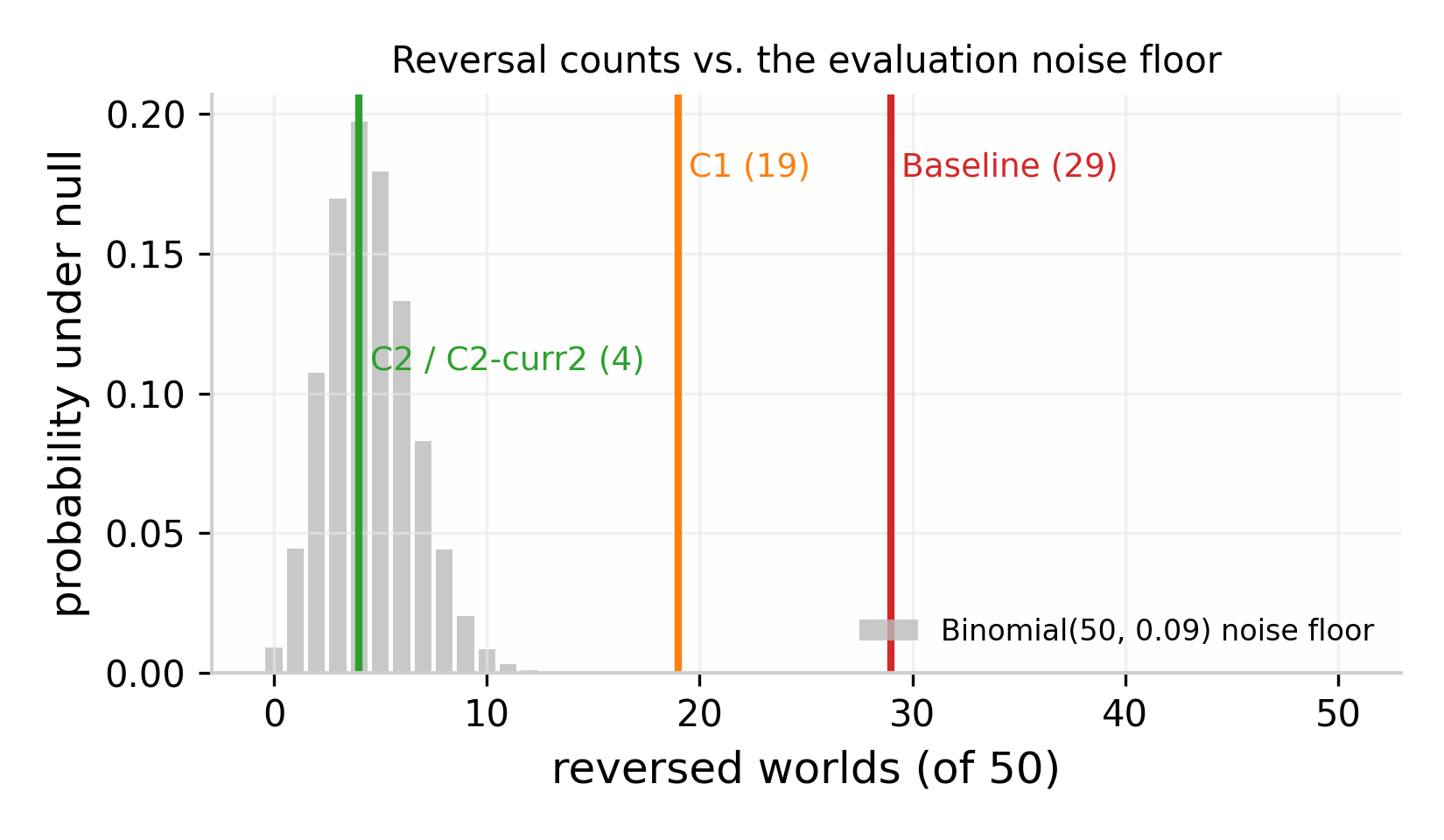}
\caption{Reversal counts vs.\ the noise floor: Binomial(50, 0.09) null with the four arms' reversal counts marked. Baseline (29) and C1 (19) lie deep in the mechanistic regime; C2 and C2-curr2 (4) sit on the null.}
\label{fig:floor}
\end{figure}

Figure~\ref{fig:floor} closes the noise-floor methodology empirically (one-sided binomial against $p_0 = 0.09$): baseline (29/50, $p = 4.7\mathrm{e}{-18}$) and C1 (19/50, $p = 2.6\mathrm{e}{-8}$) lie deep in the mechanistic regime, while C2 and C2-curr2 (4/50 each, $p = 0.67$) sit on the null---the residual reversals on the C2 side \emph{are} the measurement noise.

\subsection{Mechanism localization: the E1 inference-time decision matrix}
\label{sec:e1}

Three zero-training-cost, inference-time interventions decide the locus of suppression. Doubling the probe dose (E1a: C1 with 4 probes instead of 2) does not lift it (4/8 hit, 2 reversed); \textbf{erasing the observational evidence immediately releases the suppressed interpolation ability} (E1b: 7/9 hit, $\rto$ from $-0.09$ to $+0.56$, matching C2 at $+0.418$ same protocol / $+0.55$ same world set); injecting observational records into C2 (E1c) collapses parsing to zero (out-of-distribution), an invalid arm set aside. \emph{The suppression of interventional evidence by the observational prior is an inference-time phenomenon}---the capability is not missing; it is switched off by context.

\subsection{Robustness: seeds, curriculum, confounding, topology}
\label{sec:robust}

\textbf{Seed replication; suppression is a world property.} C1 closes on three seeds with nearly identical profiles (s0: 18/13/19; s1: 18/13/18; s2: 18/13/19) and C2 on two (s0: 41/5/4; s1: 44/3/3; McNemar $p = 0.549$). A matched-protocol second C1 seed (identical evaluation code, the same 50 worlds, $n_{rep} = 8$) replicates the per-world structure: per-world slopes correlate at $r = +0.825$, the reversed sets overlap in 14 of 18 (hypergeometric $p = 8.7\mathrm{e}{-6}$), and \emph{every} measurable strong reversal persists (11/11 at slope $< -0.15$; one world lost to a parse failure), while the worlds flowing in or out of the reversed set all sit within $|\mathrm{slope}| < 0.15$ of the threshold---boundary noise at the measurement floor, not structural drift (s0$\leftrightarrow$s2: $+0.872$, 15/19; Appendix~\ref{app:overlap}).

\textbf{Curriculum and confounding-strength transfer.} C2-curr2 (harder curriculum) is indistinguishable from C2 (hits 40 vs.\ 41, McNemar $p = 1.00$; $r = +0.775$), and C2-rho06---trained entirely at the milder $\rho = 0.6$, evaluated at $\rho = 0.9$---achieves 39/8/3, mean $\rto$ $+0.348$ (McNemar $p = 0.727$; $r = +0.684$).

\textbf{Within-run degradation.} The C1-s2 trajectory replays \S\ref{sec:h1} within a single run: the slope ratio falls from $+0.172$ to $-0.266$ over the four evaluations after step 12k while $\Delta$ stays positive.

\textbf{Novel topology.} On the unseen 13--16-node domain (C2 checkpoint), 16 of 20 worlds parse, yielding 12 hit / 2 flat / 2 reversed, $\Delta = +0.381$; the 2 reversals are floor-consistent ($p = 0.43$): causal-direction interpolation \textbf{transfers to larger novel topologies}.

\subsection{Scaling to 1B: the suppression persists as a rate}
\label{sec:e6}

\begin{table}[htbp]
\caption{Scale comparison on the identical 50-world protocol (in-distribution context length; C2 25M as in Table~\ref{tab:main}). Absolute gains shrink globally at 1B, yet the arm contrast in reversals survives. parse = fraction of syntactically valid responses.}
\label{tab:scale}
\centering\small
\begin{tabular}{llcccccc}
\toprule
Arm & Scale & Hit & Flat & Reversed & mean $\rto$ & $\Delta$ & parse \\
\midrule
C2 & 1B & 14/50 & 33 & 3 & $+0.072$ & $+0.162$ & 1.00 \\
C1 & 25M & 18/50 & 13 & 19 & $-0.092$ & $+0.041$ & 1.00 \\
C1 & 1B & 12/44 & 18 & 14 & $+0.008$ & $-0.003$ & 0.88 \\
\bottomrule
\end{tabular}
\end{table}

We retrain both arms at 0.93B (same recipe, lr $3\mathrm{e}{-4}$, 20k steps) and evaluate on the identical 50-world set (Table~\ref{tab:scale}). \textbf{(i) The arm contrast is scale-robust.} At 1B the mixed-context arm still reverses 14/44 worlds (31.8\%) against 3/50 (6\%) in the matched probe-only arm (Fisher exact $p = 1.2\mathrm{e}{-3}$). \textbf{(ii) Absolute magnitudes shrink globally and cannot measure the suppression.} Mean $|\rto|$ drops from 0.464 to 0.105 on C2 ($4.4\times$) and from 0.372 to 0.160 on C1 ($2.3\times$)---a global gain contraction that hits the control arm \emph{harder} (its origin open, \S\ref{sec:limits}); only scale-free measures (reversal rates, the arm contrast) are meaningful cross-scale, and they show persistence. \textbf{(iii) The identity of suppressed worlds does not transfer across scale.} Of the 17 worlds reversed by C1 at 25M, only 6 remain reversed at 1B (base rate 5.4), and per-world slopes are rank-uncorrelated across scale (Spearman $r = +0.08$, $n = 44$ paired)---against $r = +0.87$ across seeds \emph{within} 25M (\S\ref{sec:robust}): at 1B the suppression is a diffuse, \emph{reshuffled} rate elevation rather than stable per-world capture (Table~\ref{tab:turnover}).

\subsection{Mechanism discrimination: a graded, content-mediated switch (E5)}
\label{sec:e5}

Two inference-time probe batteries on the C1 checkpoint ($n_{probe} = 4$, $n_{rep} = 8$, 50 worlds; paired on commonly parsed subsets).

\textbf{Dose titration} (\textsc{obs} records $n_{obs} = 1/2/4$; paired $n = 37$): hits 31 $\to$ 22 $\to$ 17 and mean $\rto$ $+0.472 \to +0.204 \to +0.037$ as the observational dose grows (McNemar n1 vs.\ n2 $p = 3.9\mathrm{e}{-3}$; n1 vs.\ n4 $p = 1.3\mathrm{e}{-3}$; Wilcoxon n1 vs.\ n4 $p = 1.0\mathrm{e}{-5}$). The \emph{shape} is informative: at $n_{obs} = 2$ the model mostly goes \textbf{flat} (12 flat / 3 reversed); at $n_{obs} = 4$ it \textbf{reverses} (11 reversed, binomial $p = 3\mathrm{e}{-4}$)---a single \textsc{obs} record barely suppresses (39/50 hits, $\approx$ the probe-only level), two dilute the signal, four capture it: \emph{first dilution, then capture}. (The $n_{obs} = 8$ arm was lost to a parse-rate collapse, 0.16; \S\ref{sec:limits}.)

\textbf{Four-state content manipulation} ($n_{obs} = 4$; paired $n = 33$): hits---\textbf{\textsc{same}} (this world's true obs) 16 $<$ \textbf{\textsc{scramble}} (marginals intact, correlations destroyed) 20 $\approx$ \textbf{\textsc{cross}} (a foreign world's intact obs) 21 $<$ \textbf{\textsc{zero}} (all-zero placeholders) 25; mean $\rto$ $+0.059/+0.203/+0.192/+0.378$. The decisive contrast is \textbf{\textsc{same} vs.\ \textsc{zero}} (McNemar $p = 0.023$; Wilcoxon $p = 2\mathrm{e}{-4}$): removing observational \emph{content} releases suppression even when obs-\emph{format} tokens remain---the pure format-dilution hypothesis is rejected. \textsc{cross} and \textsc{scramble} are indistinguishable (Wilcoxon $p = 0.87$): real-valued observational content suppresses partially \emph{regardless of correlational coherence} with the current world; \textsc{same} adds the largest increment (vs.\ cross/scramble: Wilcoxon $p \approx 0.07$--$0.08$, marginal at $n = 33$). Only \textsc{same} shows mechanistic reversals (9/33; the rest noise-compatible).

\textbf{Verdict.} The switch is \emph{content-mediated and graded}, not a binary format gate: format without content (\textsc{zero}) $\approx$ full release; plausible observational numerics alone (\textsc{cross} $\approx$ \textsc{scramble}) yield partial, correlation-agnostic suppression; the world's own correlation contributes the decisive increment. Each observational record shifts the evidence mixture toward the observational attractor, and coherent this-world correlations shift it most.

\subsection{External audit on CLadder: a sign prior and one-sided reading}
\label{sec:e7}

To stress-test external validity we structurally transplant CLadder v1 \citep{jin2023cladder} into our symbolic channel: its binary Bernoulli SCMs (10 textbook graph families, 3--4 nodes, fully observed CPTs) are instantiated as our worlds and evaluated with the \emph{identical} probe protocol ($n_{probe} = 4$, $n_{rep} = 8$; $x_0 \in \{0, 1\}$, so the two-point slope is the ATE). Ground truth, recomputed by exact enumeration of the released CPTs, agrees on 9/10 families ($< 10^{-9}$; exception in Appendix). Two suites: \textsc{all} (5 worlds per family, 45) and \textsc{conf} (50 \emph{confounding}-family worlds---the direct analog of our $\rho = 0.9$ Simpson design).

\begin{figure}[!htb]
\centering
\includegraphics[width=0.55\textwidth]{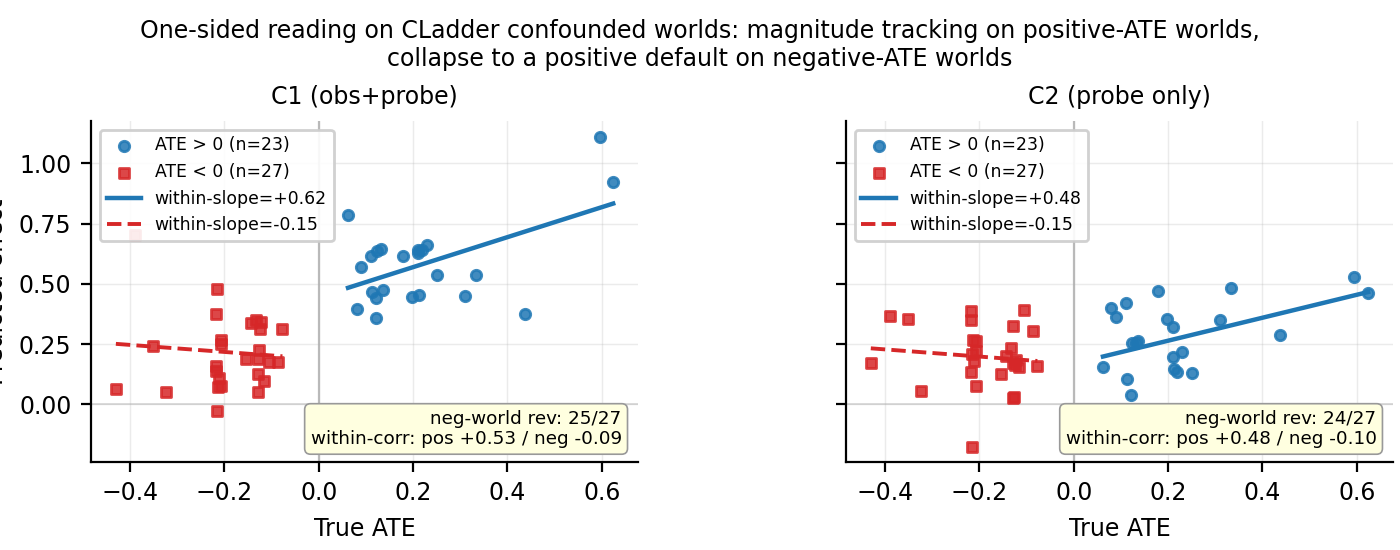}
\caption{One-sided reading on the 50 CLadder confounding worlds. Blue circles: positive-ATE worlds ($n = 23$)---both arms track magnitude (within-subset slopes $+0.62$ / $+0.48$). Red squares: negative-ATE worlds ($n = 27$)---predictions collapse onto a positive default (slopes $\approx 0$; reversals 25/27 C1, 24/27 C2). The pooled-slope difference ($a = 0.76$ vs.\ $0.21$) is positive-world amplitude, not sign reading: only 2--4\% of predictions are negative.}
\label{fig:offsetgain}
\end{figure}

Three findings. \textbf{(i) Format transfer is perfect}: 285/285 worlds parse across arms and suites---binary values, unseen graph families, and smaller graphs pose no format problem. \textbf{(ii) Both interventional arms carry a positive response offset} while the observational baseline is flat ($a \approx 0.03$, $b \approx 0$, and 47--72\% of its predictions are negative): regressing predicted on true effects, $\widehat{\Delta} = a\cdot\mathrm{ATE} + b$, gives $b = +0.233$ for C2 (95\% CI $[+0.195, +0.269]$) and $b = +0.386$ for C1 ($[+0.341, +0.431]$). The offset is therefore \emph{learned from interventional training with single-signed effects}, not a property of the evaluation pipeline. Because every training world had a positive causal effect by construction, the models answer ``$\mathrm{do}(X)$ raises $Y$'' regardless of evidence: on sign-balanced CLadder this yields ${\sim}50\%$ hit / ${\sim}50\%$ reversal (96.3\% of negative-ATE predictions remain positive). This is a genuine prior, not a measurement limit: an oracle reader of the same evidence misreads the sign only 19--45\% of the time at these effect sizes (sign-information floor, Appendix). \textbf{(iii) Both arms read \emph{one-sidedly}} (Figure~\ref{fig:offsetgain}): within positive-ATE worlds ($n = 23$) both track magnitude (within-subset correlations $+0.53$ C1 / $+0.48$ C2, C1 at ${\sim}2\times$ the amplitude, mean prediction $+0.58$ vs.\ $+0.28$); on negative-ATE worlds ($n = 27$) \emph{both} collapse onto the positive default (within-subset correlations $-0.08$ / $-0.10$; reversals 25/27 and 24/27). C1's larger pooled slope ($a = 0.76$ vs.\ $0.21$) is thus an \emph{amplitude} effect on positive worlds, not sign competence: only 2--4\% of all predictions are negative at all. The mixed context---suppressed in-distribution---amplifies magnitude tracking out-of-distribution, but neither arm reads the sign. A matched-protocol second C1 seed replicates the asymmetry but not its amplitude (positive-world correlation $+0.19$; negative-world reversals 13/27): the one-sided \emph{structure} is seed-robust, the magnitude is not (\S\ref{sec:e8}).

The audit thus exposes a boundary condition invisible to our main experiments (single-signed effects by design), quantified as an offset--gain decomposition rather than an accuracy score; it motivates the sign-randomized retraining (confounding path co-flipped, so the Simpson property holds for both signs) whose falsifiable prediction---$b \to 0$ with $a$ intact---is tested in \S\ref{sec:e8}.

\subsection{Sign-randomized retraining: the positive prior is two-layered}
\label{sec:e8}

The audit of \S\ref{sec:e7} motivates a direct repair: retrain with the causal-effect sign randomized per world ($s \in \{\pm 1\}$ multiplying \emph{both} the causal and the confounding path, so the Simpson property holds for both signs and ``always positive'' is no longer learnable). Two arms (C1-sr, C2-sr), same 25M recipe, then the identical CLadder re-test. The pre-registered prediction was $b \to 0$ with $a$ intact.

\textbf{In-distribution, the repair works---for the probe-only arm.} C2-sr reaches slope ratio $+0.74$ on the training-distribution probe battery (vs.\ $+0.18$ non-sr): forced to read the sign from interventional evidence, the model does. C1-sr flatlines at $-0.06$: with observational records present, probe learning is \emph{diluted to nothing}---the dilution endpoint of the E5 dose curve (\S\ref{sec:e5}), not capture, since no gain ever materializes to be captured.

\textbf{Out-of-distribution, the prediction is falsified.} On CLadder the offset does not move: $b = +0.21 \pm 0.02$ across all four arm$\times$suite cells, indistinguishable from the non-sr arms ($+0.233$). The gain tells a split story. C2-sr carries genuine ATE signal that the \emph{flip-the-observation} heuristic cannot explain: partial $\Corr(\widehat{\Delta}, \mathrm{ATE} \mid \mathrm{obs}) = +0.37$ against the heuristic's ceiling of $+0.17$ on this suite ($\Corr(\mathrm{ATE}, \mathrm{obs}) = -0.17$); the offset is no query-grid artifact either---on a symmetric grid $\{-1, +1\}$ it persists ($b = +0.36$ C2-sr / $+0.33$ C2; mean ratio $+0.616$ / $-0.184$)---while the apparent negative-world sign-reading gain (11\%$\to$30\% at grid $\{0, 1\}$) does not (negative predictions vanish), so we do not claim it. C1-sr collapses onto the observational sign (partial $\Corr(\widehat{\Delta}, \mathrm{obs} \mid \mathrm{ATE}) = +0.66$; mean $\rto$ $-0.66$).

\textbf{Verdict: the prior is two-layered.} An \emph{in-distribution layer} (``effects are positive'') is removed by sign randomization, though the unmasked sign competence transfers only partially; an \emph{out-of-distribution default} (``answer positive when the evidence machinery fails to engage'') is immune to the same fix. Since the pure-observational baseline shows no offset on the identical pipeline ($b \approx 0$; \S\ref{sec:e7}), the default is induced by interventional training itself; its micro-mechanism remains open (\S\ref{sec:limits}).

\section{Discussion}
\label{sec:discussion}

\textbf{Mechanism statement.} (1) Interventional evidence is \emph{sufficient} to teach causal-direction interpolation (C2: 41/50, true reversal rate $\approx 0$). (2) The observational-correlation prior \emph{suppresses} the use of interventional evidence at inference time (C1: 18/50, 19 reversed), independently of probe dose (E1a). (3) The suppression is an \emph{inference-time} phenomenon: erasing observational evidence immediately releases the suppressed ability (E1b: $+0.56$). (4) The shortcut is \emph{distributionally rational}---the observational correlation is correct for most random queries (confounded pairs are only $\sim 1/n$)---and deepens with training. (5) Training-distribution priors are \emph{two-layered}: the learnable in-distribution regularity (``effects are positive'') can be removed by sign randomization, but an out-of-distribution default persists underneath it (\S\ref{sec:e8})---repairing training statistics does not repair deployment-time fallback. (6) Activation patching localizes the switch to the observational-record (obs) rows: transplanting zero-context activations at the obs rows of layers 2--3 releases the suppressed slope ($R \approx +1.0$; $+0.96$ on a second seed), non-obs rows and the final layer release little, and the effect resists sparse head-level decomposition (Appendix~\ref{app:mech}). In one sentence: \emph{the capability lives in the weights; the switch lives in the context.}

\textbf{Implications for pretraining data practice.} Mixture-ratio optimization assumes capability scales with its signal; we show the bottleneck lies instead in the \emph{evidence ecology of the inference-time context}: interventional data remains necessary (C2 proves sufficiency), yet observational content at deployment can switch the learned ability off (E5)---motivating context engineering and evidence-type gating.

\section{Limitations}
\label{sec:limits}

(1) \textbf{Fully synthetic}: a single SCM family; the CLadder audit (\S\ref{sec:e7}) covers binary textbook graphs, not natural language. (2) \textbf{Gain contraction at scale}: absolute interventional gains shrink $2$--$4\times$ at 1B in both arms; the suppression survives only as a reversal-rate elevation, its cause (budget vs.\ recipe) unseparated (\S\ref{sec:e6}); the 1B checkpoint retains format fragility (parse 0.88). (3) \textbf{The OOD positive default is mechanism-unknown} (\S\ref{sec:e8}): sign randomization removes the in-distribution prior layer but not the out-of-distribution default; why it is \emph{positive} is unidentified, and C2-sr's apparent negative-world improvement is grid-fragile. (4) \textbf{Cross-scale identity of suppressed worlds}: stable across seeds at 25M ($r = +0.83$, matched protocol; strong reversals 11/11) but not across scale (\S\ref{sec:e6}). (5) \textbf{Dose--effect unconfirmed}: the 10-world ``suppression $\propto |\text{observational correlation}|$'' did not replicate at 50 ($p = 0.14$). (6) \textbf{Format stress and seed variance}: parsing degrades at extended contexts ($n_{obs} = 8$: 0.16; four-state: 0.74--0.86); C1's $\Delta$ varies $\sim 0.5$ across seeds---per-world claims defer to the multi-seed table.

\section*{AI Use Statement}
A large-language-model assistant was used to help draft and copy-edit the manuscript, to adversarially proof-check its claims against the underlying experimental data, and to develop and debug the evaluation and patching code. All scientific claims, experimental design, data, analyses, and conclusions were made and verified by the authors, who take full responsibility for the content of this paper.

\section*{Reproducibility Statement}
All evaluation artifacts, code fingerprints (md5), and protocol notes are documented in Appendix~\ref{app:repro}; the activation-patching experiments follow preregistered protocols and decision logs summarized in Appendix~\ref{app:mech}.

\bibliography{references}
\bibliographystyle{iclr2026_conference}

\appendix
\setlength{\textfloatsep}{8pt plus 2pt minus 2pt}
\setlength{\floatsep}{8pt plus 2pt minus 2pt}
\section{A Visual Tour of the Simpson Construction}
\label{app:construction}

\begin{figure}[!htb]
\centering
\includegraphics[width=\textwidth]{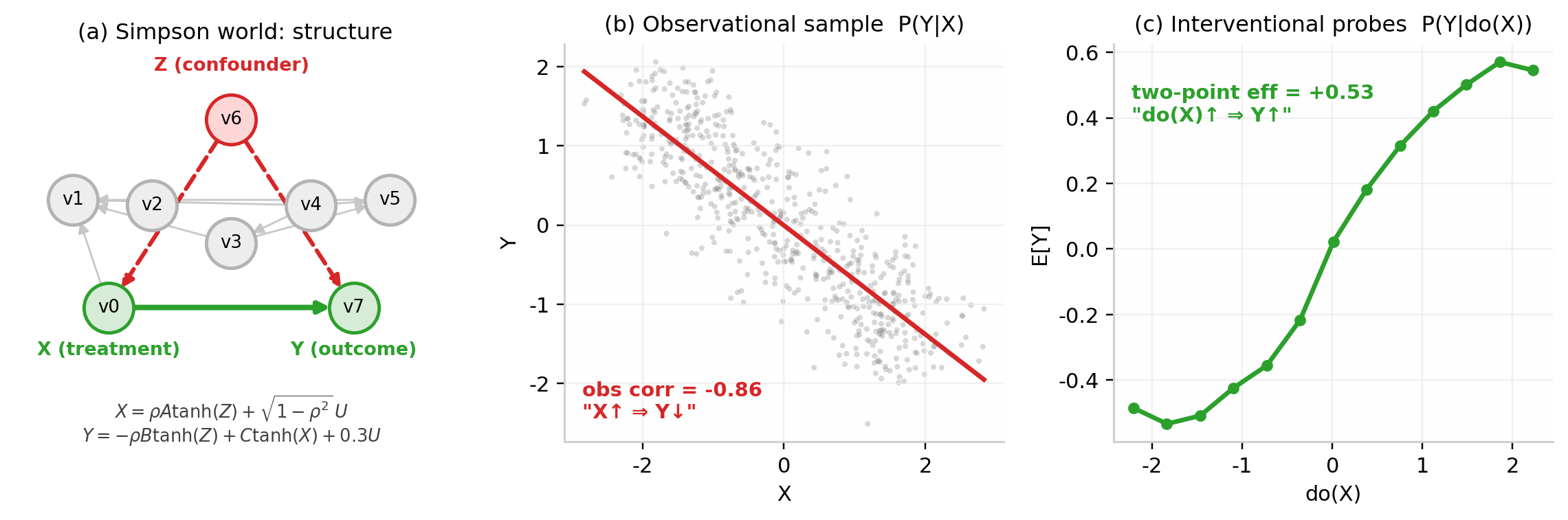}
\caption{One sampled evaluation world ($\rho = 0.9$), end to end. \textbf{(a)} Structure: the generator injects a confounded triple $(Z, X, Y)$ with $X = \rho A\tanh(Z) + \sqrt{1-\rho^2}\,U$ and $Y = -\rho B\tanh(Z) + C\tanh(X) + 0.3U$; red dashed edges are the confounding paths, the green edge is the causal path. \textbf{(b)} An observational sample from the \emph{same} world shows a strong negative trend (corr $= -0.86$). \textbf{(c)} Interventional probes $do(X)$ on the same world reveal the positive causal effect (two-point estimate $+0.53$). The sign conflict between (b) and (c) is the Simpson trap of \eqref{eq:trap}: a reader that copies observation answers ``does $X$ raise $Y$?'' backwards.}
\label{fig:construction}
\end{figure}

\section{Reproducibility Checklist}
\label{app:repro}

\textbf{Code fingerprints (md5).} \texttt{eval.py} = \texttt{e9c347704ef8ddf9524c63ddb06bf094}; \texttt{eval\_ckpt.py} = \texttt{f877ac2b3e7a02c3d910f29b7df69bde}; \texttt{mech\_patch.py} = \texttt{542e2dd7d0e88081c6b8758bb63cb8c0} (Appendix~\ref{app:mech}). All final evaluations use these versions with $n_{rep} = 8$ ($n_{rep} = 4$ for patching).

\textbf{Seed domains.} Train/val/test/novel-topology world seeds are pairwise disjoint. The 50 test worlds are \texttt{make\_world(SEED\_TEST + i, $\rho$ = 0.9, curriculum = 1, $n \in [5,12]$)} for $i = 0, \dots, 49$; world identities (hashes) are logged in each evaluation artifact.

\textbf{Cache caveat.} The generator's intervention-value cache persists across calls; offline replay of the evaluation stream must re-seed it explicitly, otherwise identical protocols drift between runs.

\textbf{Evaluation artifacts.} Every number in this paper traces to a logged evaluation file containing the checkpoint path, protocol flags, the corrected hit summary, and the per-world eff/slope/ratio/$\rto$ lines (50 worlds $\times$ 8 final arms, plus E1/E5 and novel-topology).

\section{Noise-Floor Simulation}
\label{app:floor}

Simulating a perfect least-squares interpolator against the oracle generator (4 aligned probe records per query $x_0$, 5-point regression grid; 20 repetitions of the 10-world protocol) gives a single-shot sign-error rate of 2.6/10 (26\%). Evidence averaging over $n_{rep} \in \{1, 2, 4, 8\}$ draws reduces $\mathrm{Var}[\tilde{\mu}]$ by $\approx 1/n_{rep}$ (floors 26\% / 17\% / 12\% / 9\%); we adopt $n_{rep} = 8$ (floor 0.9/10) for all final results.

\section{Cross-Seed Reversed-World Overlap}
\label{app:overlap}

Across C1 seeds the reversed-world sets overlap far beyond chance: s0$\leftrightarrow$s2 in 15/19 (hypergeometric $p = 4.2\mathrm{e}{-6}$; per-world $\rto$ Pearson $r = +0.872$, $p = 1.8\mathrm{e}{-16}$). The matched-protocol s0$\leftrightarrow$s1 overlap (14/18; strong reversals 11/11; boundary flow within $|\mathrm{slope}| < 0.15$) is reported in \S\ref{sec:robust}.

\textbf{Cross-scale turnover} (25M $\to$ 1B; Table~\ref{tab:turnover}). The reversed set turns over symmetrically (11 out / 8 in; McNemar $p = 0.65$), per-world slopes are rank-uncorrelated (Spearman $r = +0.084$, $p = 0.59$, $n = 44$; cross-seed $r = +0.87$), and survival is non-monotone in 25M strength---reshuffling, not attenuation under gain contraction.

\begin{table}[htbp]
\caption{Cross-scale turnover of the C1 reversed set (25M $\to$ 1B, identical 50-world protocol; classes by the $\pm 0.05$ slope dead zone).}
\label{tab:turnover}
\centering\scriptsize
\begin{tabular}{lll}
\toprule
Fate of the 19 reversed-at-25M worlds & $n$ & Detail \\
\midrule
Survived (rev $\to$ rev) & 6 & strength-scattered (ranks 2--19/19); weakest ($-0.057$) persists \\
Lost to flat & 7 & incl.\ $-0.564 \to +0.027$,\; $-0.398 \to -0.026$ \\
Flipped to hit & 4 & incl.\ $-0.398 \to +0.172$,\; $-0.069 \to +0.337$ \\
Lost to parse failure & 2 & incl.\ strongest reversal ($-0.681$); Fisher $p = 1.0$ \\
New reversals at 1B (from non-rev) & 8 & 4 from flat, 4 from hit (incl.\ $+0.293/+0.318 \to$ rev) \\
\bottomrule
\end{tabular}
\end{table}

\section{E5 Paired Tables}
\label{app:e5}

\begin{table}[htbp]
\caption{\textbf{(a)} E5 dose titration on the C1 checkpoint (paired $n = 37$; $n_{probe} = 4$, $n_{rep} = 8$). \textbf{(b)} E5 four-state content manipulation (paired $n = 33$; $n_{obs} = 4$; states as defined in \S\ref{sec:e5}).}
\label{tab:e5tit}
\centering\scriptsize
\begin{minipage}[t]{0.34\linewidth}
\centering
\begin{tabular}[t]{lcccc}
\toprule
$n_{obs}$ & Hit & Flat & Rev & mean $\rto$ \\
\midrule
1 & 31 & 5 & 1 & $+0.472$ \\
2 & 22 & 12 & 3 & $+0.204$ \\
4 & 17 & 9 & 11 & $+0.037$ \\
\bottomrule
\end{tabular}
\end{minipage}\hfill
\begin{minipage}[t]{0.63\linewidth}
\centering
\begin{tabular}[t]{lcccc}
\toprule
State & Hit & Flat & Rev & mean $\rto$ \\
\midrule
\textsc{same} & 16 & 8 & 9 & $+0.059$ \\
\textsc{cross} & 21 & 6 & 6 & $+0.192$ \\
\textsc{scramble} & 20 & 8 & 5 & $+0.203$ \\
\textsc{zero} & 25 & 7 & 1 & $+0.378$ \\
\bottomrule
\end{tabular}
\end{minipage}
\end{table}

\section{Protocol-Transparency Note: Context-Length Mismatch in the C1 Arm}
\label{app:protocol}

The C1 training pipeline constructs \textsc{obs\_probe} contexts from 4 observational records $+$ 2 probe records ($n_{probe}$ defaults to 2), i.e.\ an 8-line format; our initial battery queried with $n_{probe} = 4$ (12 lines)---an out-of-distribution extrapolation for C1 (C2 is unaffected: its probe branch always uses the training context length); we therefore report C1 under the in-distribution $n_{probe} = 2$ protocol throughout (\S\ref{sec:main}, \S\ref{sec:e6}).

Two consequences. \textbf{(i) The main result is protocol-robust}: re-evaluating C1 at 25M under $n_{probe} = 2$ reproduces the out-of-distribution battery exactly (18/13/19, mean $\rto$ $-0.092$, $\Delta$ $+0.041$)---the 12-line evaluation did not inflate suppression, and probe-dose doubling $2 \to 4$ provides zero rescue under observational evidence (a dose dual to \S\ref{sec:e5}). \textbf{(ii) Scale-dependent format fragility}: the 1B C1 checkpoint fails to extrapolate to 12 lines (parse 0.32; missing variables, mis-closed tags) and stays fragile even in-distribution (parse 0.88), while 25M parses perfectly under both---a scale-by-recipe robustness issue (lr $3\mathrm{e}{-4}$ vs.\ $6\mathrm{e}{-4}$), not a causal finding; all 1B conclusions use the in-distribution protocol.

\section{Negative-Result Registry}
\label{app:neg}

\textbf{E1c (C2 + obs injection)}: parse rate 0 (obs format out-of-distribution for a probe-only model); arm invalid, set aside. \textbf{Dose--effect non-replication}: ``suppression $\propto |\Corr_{obs}|$'' (10 worlds) fails at 50 ($|r|$ 0.770 rev vs.\ 0.729 hit, $p = 0.14$; \S\ref{sec:limits}). \textbf{Ablation-A target-node confusion}: rejected ($|\eff(Z{\to}Y)|/|\eff(X{\to}Y)|$ \emph{lower} in the reversed group). \textbf{$n_{obs} = 8$ titration arm}: lost to parse-rate collapse (0.16; long-context format stress).

\section{Mechanism Report: Activation-Patching Localization}
\label{app:mech}

\textbf{Protocol} (preregistered; each stage's gates were fixed before that stage's data, with one documented revision after the engine self-test and before any downstream data). For each reversed world we measure paired baselines---the $do()$-slope under the same-world obs context ($s_{\mathrm{same}}$, suppressed) and under zeroed obs ($s_{\mathrm{zero}}$, released)---then transplant activations from the zero-condition run into the same-condition run at a candidate site and score the release ratio $R = (s_{\mathrm{p}} - s_{\mathrm{same}})/(s_{\mathrm{zero}} - s_{\mathrm{same}})$ ($R{=}1$: full release; worlds with $s_{\mathrm{zero}} - s_{\mathrm{same}} \le 0.05$ excluded by guard). Sites: residual-stream obs-record vs.\ non-obs positions per layer (G1); individual heads/MLPs selected on dev, verified on holdout (G2). $n_{rep} = 4$; reversed worlds from the formal 50-world archive (s0: 19, deterministic dev/holdout split; s1: all 18, no split).

\begin{figure}[H]
\centering
\includegraphics[width=\textwidth]{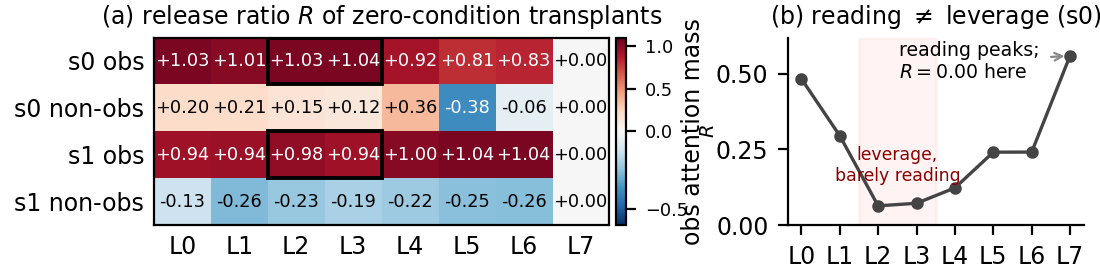}
\caption{\textbf{(a)} Release ratio $R$ of zero-condition activation transplants, by layer and row type (s0 dev / s1 all-reversed; $n_{rep} = 4$). The switch travels with the obs rows ($R \approx +1.0$ from L0 through L6, both seeds) and vanishes exactly at L7; black boxes mark the preregistered L2/L3 obs cells. \textbf{(b)} Obs-directed attention mass by layer (s0): reading peaks at L7, where $R = 0.00$---reading $\neq$ leverage.}
\label{fig:mech}
\end{figure}

\textbf{Gate results} (two seeds; Figure~\ref{fig:mech}a). \textbf{G1 layer scan} (s0 dev, $n = 8/10$): obs rows L0--L3 $R = +1.01$--$+1.04$, L4--L6 $0.81$--$0.92$, L7 exactly $0.000$; non-obs rows $\le +0.21$ at L0--L3 ($+0.36$ max) $\Rightarrow$ PASS. \textbf{G2 component circuit} (s0 holdout, $n = 8/9$): dev-selected top-8 union $R = +0.20$ (CI95 $[-0.43, +0.67]$; vs.\ random $\Delta = +0.25$, permutation $p = 0.19$); reverse patch $R_{\mathrm{rev}} = +0.14$ $\Rightarrow$ FAIL: no sparse circuit isolated under this search. \textbf{G1-s1 cross-seed confirm} (s1 all rev, $n = 17/18$): L2/L3 obs $R = +0.98/+0.94$ (L0--L6 obs rows all $+0.94$ to $+1.04$); obs $-$ non-obs $\Delta = +1.20/+1.13$; non-obs all negative ($-0.13$ to $-0.26$); L7 exactly $0.000$ $\Rightarrow$ PASS.

\textbf{Controls and reading.} Component-stage patches are surgically clean (KL sentinel max $2.9\mathrm{e}{-7}$ nats) and leave hit worlds intact (4/5 signs, mean $|\Delta\mathrm{slope}| = 0.15$); a random 3-head baseline at L3 obs reaches $|R| \approx 0.4$, calibrating the G2 null. Obs-directed attention mass is U-shaped (L0 $0.484 \to$ L2/L3 $0.064/0.072 \to$ L7 $0.560$) but leverage is not: it saturates from L0 through the barely-reading middle layers and vanishes exactly at L7, where reading peaks (Figure~\ref{fig:mech}b). We therefore claim a \emph{row-level, distributed} switch and \emph{sufficiency only}---these data do not speak to necessity.

\end{document}